\documentclass[sigconf]{acmart}
\AtBeginDocument{%
  \providecommand\BibTeX{{%
    \normalfont B\kern-0.5em{\scshape i\kern-0.25em b}\kern-0.8em\TeX}}}

\author{Ming Shan Hee}
\affiliation{%
  \institution{Singapore University of \\ Technology and Design}
  \city{Singapore}
  \country{Singapore}
}
\email{mingshan_hee@mymail.sutd.edu.sg}

\author{Aditi Kumaresan}
\affiliation{%
  \institution{Singapore University of \\ Technology and Design}
  \city{Singapore}
  \country{Singapore}
}
\email{aditi_kumaresan@mymail.sutd.edu.sg}

\author{Nguyen Khoi Hoang}
\affiliation{%
  \institution{VinUniversity}
  \city{Hanoi}
  \country{Vietnam}
}
\email{20nguyen.hk@vinuni.edu.vn}

\author{Nirmalendu Prakash}
\affiliation{%
  \institution{Singapore University of \\ Technology and Design}
  \city{Singapore}
  \country{Singapore}
}
\email{nirmalendu_prakash@sutd.edu.sg}

\author{Rui Cao}
\affiliation{%
  \institution{Singapore Management University}
  \city{Singapore}
  \country{Singapore}
}
\email{ruicao.2020@phdcs.smu.edu.sg}

\author{Roy Ka-Wei Lee}
\affiliation{%
  \institution{Singapore University of \\ Technology and Design}
  \city{Singapore}
  \country{Singapore}
}
\email{roy_lee@sutd.edu.sg}

\renewcommand{\shortauthors}{Ming Shan Hee et al.}

\copyrightyear{2023} 
\acmYear{2023} 
\setcopyright{acmlicensed}\acmConference[MM '23]{Proceedings of the 31st ACM International Conference on Multimedia}{October 29-November 3, 2023}{Ottawa, ON, Canada}
\acmBooktitle{Proceedings of the 31st ACM International Conference on Multimedia (MM '23), October 29-November 3, 2023, Ottawa, ON, Canada}
\acmPrice{15.00}
\acmDOI{10.1145/3581783.3613463}
\acmISBN{979-8-4007-0108-5/23/10}

\usepackage{multirow}
\usepackage{bm}

\usepackage[belowskip=2pt,aboveskip=0pt]{caption}

\begin{document}
\title{MATK: The Meme Analytical Tool Kit}



\begin{abstract}
The rise of social media platforms has brought about a new digital culture called memes. Memes, which combine visuals and text, can strongly influence public opinions on social and cultural issues. As a result, people have become interested in categorizing memes, leading to the development of various datasets and multimodal models that show promising results in this field. However, there is currently a lack of a single library that allows for the reproduction, evaluation, and comparison of these models using fair benchmarks and settings. To fill this gap, we introduce the Meme Analytical Tool Kit (MATK), an open-source toolkit specifically designed to support existing memes datasets and cutting-edge multimodal models. MATK aims to assist researchers and engineers in training and reproducing these multimodal models for meme classification tasks, while also providing analysis techniques to gain insights into their strengths and weaknesses. To access MATK, please visit \url{https://github.com/Social-AI-Studio/MATK}.


\end{abstract}




\maketitle

\section{Introduction}

Memes have become a crucial means of communication and cultural expression in recent years, influencing our digital interactions and content consumption. These images or videos, often accompanied by clever captions, spread rapidly across the internet, serving as a medium for conveying complex ideas, humor, and social commentary. As a result, this popular digital phenomenon has attracted attention from both industry and academia, leading to the analysis of its underlying topics and impact. Notably, Facebook launched "The Hateful Meme Challenge," releasing a dataset of over 10,000 memes annotated for hateful and non-hateful attributes~\cite{kiela2020hateful}. Subsequently, further studies have emerged, exploring various applications of meme analysis, such as offensive content detection~\cite{suryawanshi2020multimodal}, sentiment analysis~\cite{sharma2020semeval}, and identification of victims and roles~\cite{fharook2022you,pramanick2021detecting}.

The availability of diverse and extensive datasets has significantly contributed to the rapid development and analysis of multimodal models for memes. These datasets serve as the foundation for training and evaluating models, enabling researchers to create more accurate and robust solutions for classifying memes across different domains. Researchers have explored different approaches, including classic two-stream models that combine text and visual features for classifying hateful memes~\cite{kiela2020hateful,suryawanshi2020multimodal}, as well as fine-tuning large-scale pre-trained multimodal models for multimodal classification tasks~\cite{muennighoff2020vilio,pramanick2021momenta,yang2022multimodal,sandulescu2020detecting,cao2023prompting,sharma2022detecting}. However, a challenge arises when models and datasets are scattered across separate GitHub repositories. With the increasing number of meme datasets and multimodal models, it becomes progressively difficult to locate and reproduce specific models on unseen datasets. This dispersion hampers research progress as substantial effort and time is required to find and adapt custom multimodal models for meme classification tasks. Therefore, establishing a centralized repository that facilitates the organization, discovery, and sharing of models and datasets can greatly enhance efficiency and foster collaborative advancements in the field of meme classification and analysis.

\begin{figure*}[ht]
\centering
\includegraphics[width=0.95\textwidth]{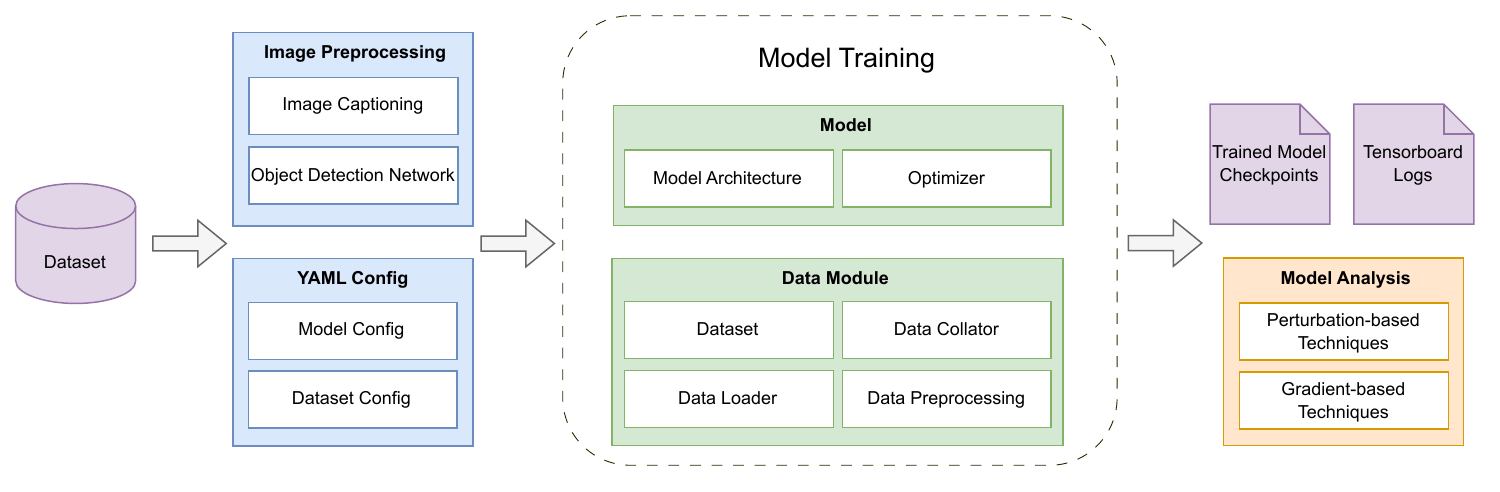}
\caption{Overall framework of MATK repository}
\label{fig:architecture-diagram}
\end{figure*}

To address this issue, we have developed the Meme Analysis Tool Kit (MATK), an open-source tool kit explicitly created to support the expanding collection of meme datasets and cutting-edge models for meme classification tasks. MATK serves as a valuable resource for machine learning practitioners and socio-linguistic researchers, simplifying the training process of state-of-the-art models. Additionally, it incorporates post-hoc model interpretation techniques such as LIME \cite{ribeiro2016should} and SHAP \cite{lundberg2017unified}, facilitating the analysis of model strengths and weaknesses. Notably, MATK encourages researchers to contribute new techniques and datasets, promoting reproducibility and fostering collaboration. The contributions of MATK can be summarized as follows:


\begin{itemize}
    \item MATK is a centralized repository that organizes, discovers, and shares meme datasets and multimodal models. It currently supports the training and analysis of seven models and eight meme datasets, with future additions planned.
    \item MATK is designed as a user-friendly tool for researchers and practitioners, especially those new to the field, to efficiently train state-of-the-art models on multiple meme datasets.
    \item MATK encourages collaborations and contributions by offering a straightforward training framework with modular components and a flexible configuration system.
\end{itemize}

\section{\textsf{MATK}: Meme Analysis Tool Kit}

In this section, we present a comprehensive overview of MATK. MATK makes use of the PyTorch Lightning framework~\footnote{\url{https://www.pytorchlightning.ai/index.html}} to simplify the training and development process for multimodal memes. It adopts an object-oriented programming approach, utilizing composition and inheritance among classes to avoid redundant implementation. Furthermore, MATK incorporates a flexible configuration system, enabling researchers to seamlessly switch between datasets and model modules. The overall framework architecture of MATK is depicted in Figure \ref{fig:architecture-diagram}.


\subsection{Design Principles}

MATK embodies user-friendliness and simplicity through the following design principles:

\textbf{Modularity:} Leveraging PyTorch Lightning, MATK allows for effortless integration of components using LightningModel and LightningDataModule. Independent modules can be developed for different tasks and input modalities, enabling seamless interfacing between new LightningDataModules and LightningModels.

\textbf{Composable Configurations:} MATK provides a user-friendly configuration system based on YAML files. These YAML files offer flexibility in specifying components and training parameters. For example, researchers can define the training parameters under the "trainer" configuration, swap the LightningModel using the "model" parameter, or swap the meme dataset using "data" parameter.

\textbf{Reproducibility:} MATK ensures reproducible implementations of multimodal models for diverse meme classification tasks. By specifying a random seed using "seed\_everything" in the YAML file, researchers and practitioners can easily reproduce results.





\subsection{Image Preprocessing}

MATK provides three sets of useful tools designed to extract meaningful visual features from memes, specifically focusing on eliminating text overlays, converting images into embedding features, and translating images into text-based features.

\textbf{Image Cleaning:} MATK offers image cleaning tools to enhance the extraction of visual features from memes. These tools effectively remove superimposed text from images without compromising the visual elements. This is achieved by using MMedit \cite{mmediting2022} and KerasOCR ~\footnote{\url{https://keras-ocr.readthedocs.io/en/latest/}} to perform inpainting on the removed text areas.

\textbf{Embedding Features:} Vision-language models (VLMs) often utilize object detection networks to extract prominent visual features from images. However, running the object detection network concurrently with the VLMs can slow down model training and increase GPU RAM usage. Hence, it is beneficial to preprocess images into embedding features before running the model. MATK provides helpful scripts for converting image modalities into embedding features. These scripts include visual feature extraction scripts that employ Faster-RCNN~\cite{ren2015faster} and CLIP~\cite{radford2021learning} models.

\textbf{Text-Based Features:} Researchers have achieved success in multimodal meme classification tasks by leveraging text-based models. One common approach to extracting and representing the semantics of an image as textual descriptions is through image captioning techniques. MATK includes image captioning scripts that utilize ClipCap~\cite{mokady2021clipcap} and BLIP-2~\cite{li2023blip} models, simplifying the process of generating image captions for text-based feature extraction.

\subsection{Data Modules}

The MATK repository incorporates various meme datasets, each encapsulated within a LightningDataModule class. This class serves the primary purpose of facilitating dataset loading, preprocessing, batch size definition, and data collation functions. The supported datasets are listed below:

\begin{itemize}
    \item \textbf{Facebook’s Hateful Memes (FHM)}~\cite{kiela2020hateful}. This dataset consists of 10,000 memes, some of which contain discriminatory content towards protected categories/classes. Each meme is labeled as hateful or non-hateful.
    \item \textbf{Facebook’s Fine-Grained Hateful Memes (FHM-FG)}~\cite{mathias2021findings}. An extension of Facebook's Hateful Memes. This dataset comprises 10,000 memes annotated with labels indicating whether they are hateful or non-hateful, along with the attack type and the targeted protected category.
    \item \textbf{Harmful Memes (HarMeme)}~\cite{sharma2020semeval}. This dataset contains 3,544 memes related to COVID-19, annotated with the intensity of harmfulness and the target class.
   \item \textbf{Multimedia Automatic Misogyny Identification}~\cite{fersini2022semeval}. This dataset consists of 10,000 memes focused on misogyny. The memes are annotated with multiple aspects of common hatred towards women, including misogyny, shaming, stereotypes, objectification, and violence.
   \item \textbf{Memotion}~\cite{sharma2020semeval}. This dataset contains 9,871 memes annotated with sentiment, type of emotion and the corresponding emotional intensity.
\end{itemize}

\subsection{Models}
MATK includes a collection of state-of-the-art models specifically designed for meme classification. These models are encapsulated within a LightningModel class, which encompasses the model's architecture and associated functions for training, validation, and testing. The following models are supported:

\begin{itemize}
    \item \textbf{BART} \cite{lewis2019bart}. A transformer encoder-decoder model that utilizes a denoising autoencoder during pre-training. It learns to reconstruct the original text by corrupting it with a customizable noising function.
    \item \textbf{T5} \cite{raffel2020exploring}. A transformer encoder-decoder model trained on a diverse range of text-to-text tasks, both supervised and unsupervised. Each supervised classification tasks are transformed into a language problems for training.
    \item \textbf{PromptHate} \cite{cao2023prompting}. An encoder model that effectively utilizes pre-trained language models by employing prompts and leveraging implicit knowledge for classifying hateful memes.
    \item \textbf{VisualBERT} \cite{li2019visualbert}. A unified single-stream vision-language model that processes both textual and visual inputs within the same module.
    \item \textbf{LXMERT} \cite{tan2019lxmert}. A two-stream vision-language model that separately processes text and images using distinct modules.
    \item \textbf{FLAVA} \cite{singh2022flava}. A foundational vision-language model pre-trained on multimodal data, including image-text pairs and unimodal data with unpaired images and text.
\end{itemize}

\subsection{Analysis Techniques}
MATK incorporates analysis techniques that allow for the assessment of the model's strengths and weaknesses. These techniques provide practitioners and researchers with valuable insights into the model's performance on unseen datasets. Currently, MATK supports the following techniques:

\begin{itemize}
    \item \textbf{LIME} \cite{ribeiro2016should}. This perturbation-based technique approximates any black box machine learning model by creating a local and interpretable model to explain individual predictions.
    \item \textbf{Integrated Gradients} \cite{sundararajan2017axiomatic}. This gradient-based technique evaluates feature importance by calculating the average of the model's output gradient interpolated along a straight-line path in the input data space.
\end{itemize}

\section{Empirical Results}
\begin{table}[t]
\centering
\small
\caption{Reproducibility Experiments on FHM. A \textbf{bolded score} indicates successful reproduction of a similar model performance, while an \underline{underlined score} denotes a considerable discrepency in model performance.}
\begin{tabular}{ccccc}
\toprule
& \multicolumn{2}{c}{MATK} & \multicolumn{2}{c}{Reported Results}  \\
\cmidrule(rl){2-3} \cmidrule(rl){4-5}
Model & Acc. & AUROC & Acc. & AUROC \\
\midrule
BART & 50.60 & 46.60  & - & -\\
T5 & 55.60 & 59.60 & - & -  \\
PromptHate & \textbf{72.45} & \textbf{81.24} & \textbf{72.98} & \textbf{81.45} \\
\midrule
VisualBERT & 61.00 & \underline{67.50} & - & \underline{74.10} \\
LXMERT & 61.20 & 66.10 & - & - \\
FLAVA & 69.40 & \textbf{77.40}  & - & \textbf{76.70} \\
\bottomrule
\end{tabular}
\label{tab:reproducibility-fhm-dataset}
\end{table}

In this section, we showcase the reproducibility and utilization of our library. We evaluated our models using the Facebook's Fine-Grained Hateful Memes dataset and performed LIME analysis on a fine-tuned FLAVA model.

\subsection{Reproducibility Experiments}
Table \ref{tab:reproducibility-fhm-dataset} shows the performance of implemented models in MATK. MATK reproduced the results of FLAVA and PromptHate models almost perfectly, with a difference of (+/- 1). However, the variation in performance of the VisualBERT model can be attributed to differences in implementation and the pre-trained model checkpoint used. MATK currently uses VisualBERT from Huggingface~\footnote{\url{https://huggingface.co/}}, while the reported results are based on VisualBERT from Facebook’s MMF~\footnote{\url{https://github.com/facebookresearch/mmf}}. This highlights the importance of a unified repository, as it ensures consistency for practitioners and researchers who might make similar assumptions. Consequently, we intend to support the model variants from Facebook’s MMF within MATK in the future.

\subsection{LIME Analysis}
\begin{figure}[t] 
	\centering
fin	\includegraphics[width=0.47\textwidth]{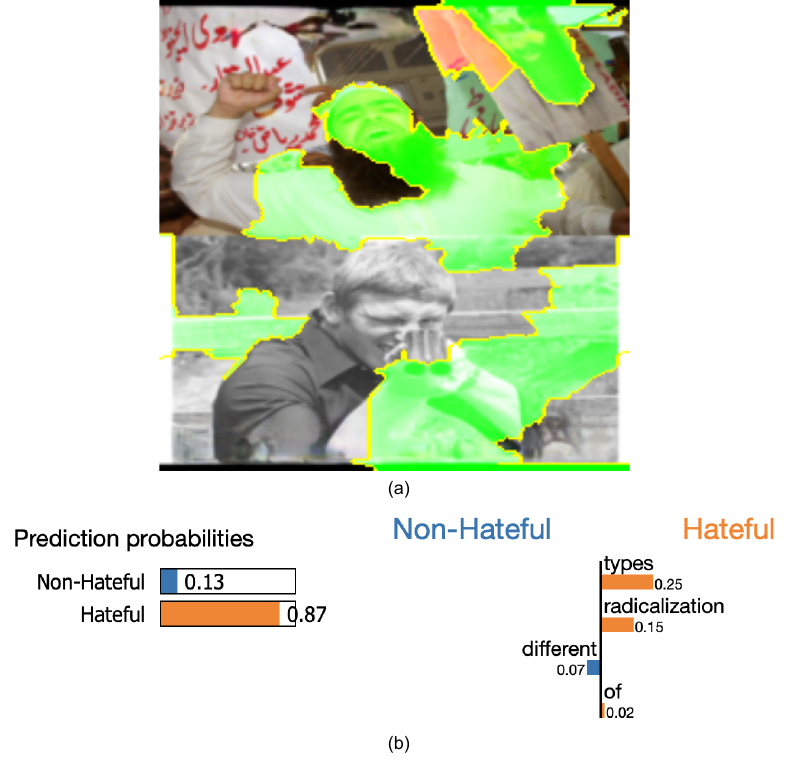}
	\caption{LIME analysis of meme image.}
	\label{fig:example}
\end{figure}

Figure \ref{fig:example} shows LIME analysis of the FLAVA model's explanation for a meme incorrectly predicted as hateful. Our observations indicate that the image's objects (Arab and riffle) and strong language ("radicalization" and "types") significantly influence the model's prediction. While this could suggest the presence of a potential bias in the trained model, we also note that the model demonstrates an understanding of negation through the negative contribution of the term "different." Consequently, a comprehensive and thorough analysis is warranted to investigate these findings further. Nonetheless, the utilization of LIME analysis provides researchers with an initial yet valuable tool for assessing model performance.


\section{Conclusion}
We present the Meme Analytic Tool Kit (MATK), a comprehensive tool designed to facilitate meme classification tasks by incorporating advanced multimodal models and supporting various meme datasets. MATK focuses on modularity, flexibility, and user-friendliness, making model training and analysis across diverse domains effortless. Importantly, MATK integrates model analysis techniques that offer practitioners and researchers valuable insights into the strengths and weaknesses of their models. Our future plans for MATK include the following: (i) expanding dataset and model support to include the TotalDefMeme dataset \cite{nirmal2023totaldefmeme}, MET-Meme dataset \cite{xu2022met}, and the DisMultiHate model \cite{lee2021disentangling}; (ii) enabling multi-task training with different datasets, such as using FHM for hateful classification and HarMeme for harmful classification; (iii) enhancing the documentation to provide comprehensive and detailed information.


\section{Acknowledgments}
This work is partially supported by the grant RS-INSUR-00027-E0901-S00.

\bibliographystyle{ACM-Reference-Format}
\balance
\bibliography{ref}

\end{document}